\documentclass[10pt,twocolumn,letterpaper]{article}

\usepackage{wacv}
\usepackage{times}
\usepackage{epsfig}
\usepackage{graphicx}
\usepackage{amsmath}
\usepackage{amssymb}
\usepackage{url}

\wacvfinalcopy 

\def\wacvPaperID{276} 

\ifwacvfinal\fi
\begin{document}

\title{TAN: Temporal Aggregation Network for Dense Multi-label Action Recognition}


\author{Xiyang Dai \space\space\space\space\space\space\space\space
Bharat Singh \space\space\space\space\space\space\space\space
	Joe Yue-Hei Ng \space\space\space\space\space\space\space\space
	Larry S. Davis\\
	Institution for Advanced Computer Studies\\\
	University of Maryland, College Park\\
	{\tt\small \{xdai, bharat, yhng, lsd\}@umiacs.umd.edu}
}

\maketitle
\ifwacvfinal\thispagestyle{empty}\fi

\begin{abstract}
We present Temporal Aggregation Network (TAN) which decomposes 3D convolutions into spatial and temporal aggregation blocks. By stacking spatial and temporal convolutions repeatedly, TAN forms a deep hierarchical representation for capturing spatio-temporal information in videos. Since we do not apply 3D convolutions in each layer but only apply temporal aggregation blocks once after each spatial downsampling layer in the network, we significantly reduce the model complexity. The use of dilated convolutions at different resolutions of the network helps in aggregating multi-scale spatio-temporal information efficiently. Experiments show that our model is well suited for dense multi-label action recognition, which is a challenging sub-topic of action recognition that requires predicting multiple action labels in each frame. We outperform state-of-the-art methods by 5\% and 3\% on the Charades and Multi-THUMOS dataset respectively.
\end{abstract}

\section{Introduction}
Convolutional Neural Networks (CNNs) have seen tremendous success across different problems including image classification \cite{krizhevsky2012imagenet, resnet, dai2017efficient}, object detection \cite{girshick2014rich, faster-rcnn, rfcn, Singh_2018_CVPR, Xu_2018_ECCV}, style transfer \cite{style, chen2017stylebank, dai2017fason, chen2018stereoscopic, fan2018decouple}, action recognition \cite{simonyan2014two, c3d, i3d, P3d} and action localization \cite{Dai_2017_ICCV, yin2018deep, zhang2018s3d, zhang2018dynamic}. In each of these problems, several application specific changes to network design have been proposed. In particular, for action recognition, a two-stream network comprising two parallel CNNs, one trained on RGB images and another trained on stacked optical flow fields, showed that incorporating temporal information into the network architecture provides a significant benefit in performance \cite{simonyan2014two}. Since optical flow computation is an additional overhead, network architectures like C3D \cite{c3d} operate only on a sequence of images and perform 3D convolutions in each layer of the network. However, 3D convolutions in each layer increase the model complexity and with just 3x3x3 convolutions, it is hard to capture larger temporal context. Therefore, we need to design a network architecture which can learn semantic representations of actions efficiently. 

\begin{figure}
    \begin{center}
        \includegraphics[width=1\linewidth]{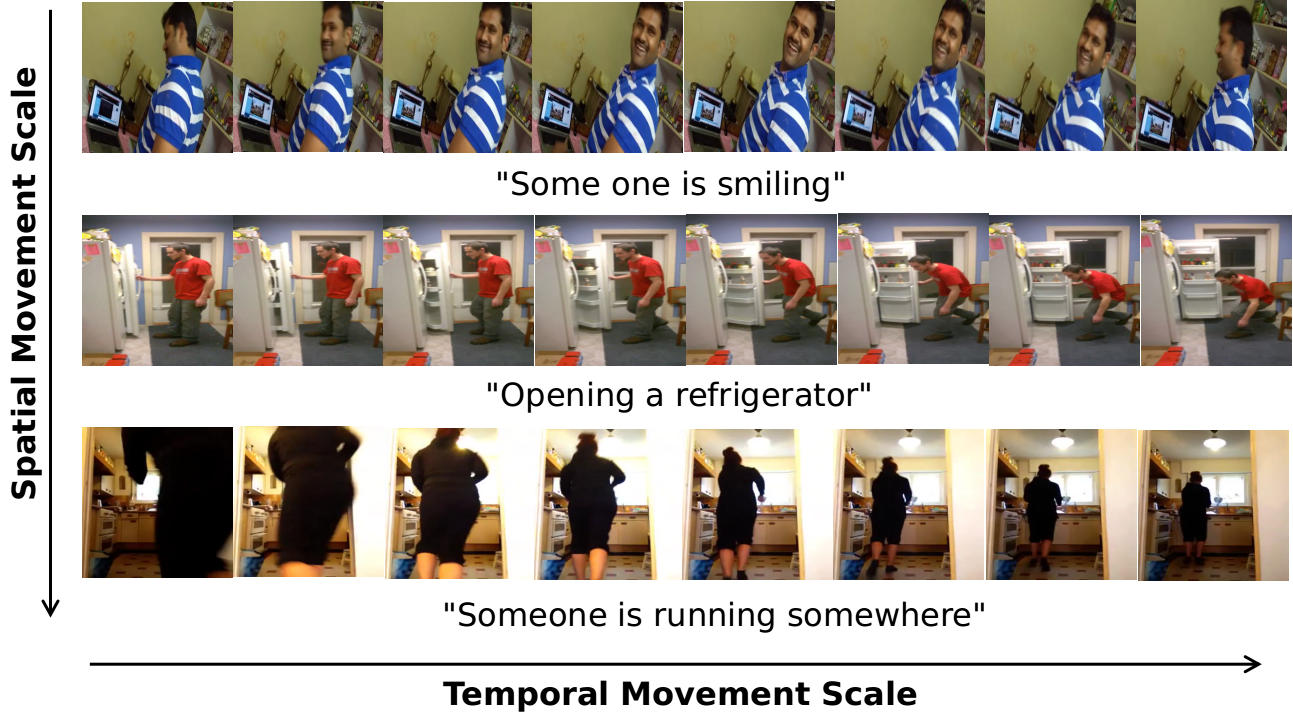}
    \end{center}
    \caption{An illustration of the spatial and temporal variances of human activities}
\label{fig:exp}
\end{figure}

While it is important to design efficient network architectures, they should also be capable of learning all the variations which appear in videos during training. 
In a video, multiple frames aggregated together represent a semantic label, so the amount of computation needed per semantic label is an order-of-magnitude larger compared to recognition tasks on images. The difficulty is further amplified because actions can span multiple temporal and spatial scales in videos. Short actions like micro-expressions (raising an eyebrow) require high-resolution spatio-temporal reasoning, while other longer actions like running or dancing involve large spatial movements, which can be identified with features with lower spatial resolution. Therefore, it is essential to aggregate multi-resolution spatio-temporal information without blowing up network complexity. 

For image recognition tasks like object detection and semantic segmentation, dilated convolutions \cite{ren2015faster,chen2016deeplab} have been widely adopted to increase receptive field sizes without increasing model complexity. By applying dilated convolutions with different filter sizes, multi-scale context can be efficiently captured. Multi-scale {\em spatio-temporal} context is important for video recognition tasks. To this end, we propose TAN, which applies multi-scale dilated temporal convolutions after every spatial downsampling layer in the network. Since the spatial receptive field of the network doubles after each downsampling layer, our network has the capacity to learn fine-grained motion patterns in the feature-maps generated closer to the input. Large scale spatio-temporal patterns are captured in deeper layers which have coarse resolution. By only applying temporal convolutions after each downsampling layer, the computational burden and model complexity are reduced when compared to methods which apply 3D convolutions in each layer of the network. The use of dilated convolutions also facilitates capturing larger temporal context efficiently. 
We conducted extensive ablation studies to verify the effectiveness of our approach.

Our architecture is especially suitable for the dense action labeling task as it offers a good balance between temporal context and bottom up visual features computed at a particular time instant. 
By modeling short-term context efficiently with convolutions, TAN obtains state-of-the-art results on two benchmark datasets, Charades and Multi-THUMOS for the dense action prediction task, outperforming existing methods by 5\% and 3\% respectively. Although TAN is designed for dense action prediction, we also applied it to action detection where it obtains state-of-the-art results, showing the effectiveness of TAN in a variety of tasks.

\section{Related Work}
\textbf{Action recognition.} Action recognition is one of the core components of video understanding. In early works, 3D motion templates \cite{bobick2001recognition}, or features such as SIFT-3D \cite{3d-sift} were used for representing temporal information for action recognition. Later, the introduction of dense trajectories \cite{dt} and improved dense trajectories \cite{idt} provided a significant boost in performance for feature based pipelines. 

After the early success of deep learning \cite{krizhevsky2012imagenet,vgg,inception} on image classification, early deep learning based video recognition methods focused on utilizing deep features. Karpathy et al. \cite{KarpathyCVPR14} evaluated different fusion methods for deep features. Wang et al. \cite{WangQT15a} pooled deep-learned features based on trajectory constraints. Simonyan et al. \cite{simonyan2014two} proposed a two-stream network which learns temporal dynamics on both stacked optical flow and appearance. Feichtenhofer et al. \cite{feichtenhofer2016convolutional} further analyzed different ways to fuse two-stream networks. Wang et al. \cite{TSN2016ECCV} further improved the two-stream network by introducing a temporal sampling strategy and training on video-level instead of short snippets to enable efficient learning on full videos.

Meanwhile, researchers started to design novel deep architectures specifically for video tasks. Tran et al. \cite{c3d} first proposed to utilize 3D convolution, which naturally extends convolutional filters to temporal domain. Then, Carreira et al. \cite{i3d} proposed to build a 3D convolution variant of inception architecture and trained on a newly collected large-scale dataset of actions. This network achieved impressive performances on multiple benchmarks but was computationally costly. Very recently, Qiu et al. \cite{P3d} proposed P3D to simplify 3D convolutions with 2D convolutional filters in the spatial domain followed by 1D convolutional filters in the temporal domain. This method is most relevant to proposed work. We also decompose 3D convolutions into separate spatial and temporal convolutions. However, unlike P3D, which replaces 3D convolution with 2D + 1D convolutions, we design a dedicated temporal aggregation block to capture context among multiple temporal levels and meanwhile preserving the temporal resolution. We only place a single layer of temporal block after each downsampling layer to reduce the filters needed for performing video recognition tasks. By stacking these two types of blocks repeatedly, we effectively capture appearance and spatio-temporal motion pattern which is important for localization tasks in videos.

\textbf{Multi-label prediction.} Multi-label dense action recognition is a considerably harder task than action recognition as it requires labeling frames with all the actions occurring in them. Yeung \cite{multithumos} first proposed this task by densely labeling every frame within a popular sports action dataset \cite{THUMOS14}. They further applied standard techniques such as LSTM, two-steam networks and created a very strong baseline. Sigurdsson et al. \cite{charades} collected another multi-label dataset by crowd-souring everyday activities at home. For recognizing these actions, Girdhar et al. \cite{Girdhar_ActionVLAD_cvpr17} proposed ActionVLAD, that aggregates appearance and motion features from two-stream networks. Sigurdsson et al. \cite{sigurdsson2017asynchronous} used a fully-connected temporal CRF model for reasoning over various aspects of activities. Dave et al. \cite{predictive_cvpr17} employed RNNs to sequentially make top-down predictions and later then corrected them by bottom-up observations. Most recently, Sigurdsson et al. \cite{Sigurdsson_what_2017_ICCV} performed a detailed analysis on what kinds of information are needed to achieve substantial gains for activity understanding among objects, verbs, intent, and sequential reasoning.

Our approach is suitable for dense multi-label action recognition because we learn dense spatio-temporal information efficiently without the need to reduce temporal resolution. Our model creates a hierarchical spatio-temporal representation that captures context among different temporal frequencies, which significantly improves performance on benchmark datasets. 

\textbf{Temporal action localization.} Unlike action recognition where we predict a label per video, action detection requires predicting temporal boundaries of an action in an untrimmed video. Recent methods mainly focus on two types of approaches: making dense predictions or predict temporal boundaries (like proposals). Escorcia et al. \cite{daps} encoded a stream of frames into a single feature vector using an LSTM which was then used to rank proposals. Singh et al. \cite{bharat} used a bi-directional LSTM on multi-stream features and performed fine-grained action detection. Shou et al. \cite{scnn_shou_wang_chang_cvpr16} utilized multiple temporal scales to rank candidate segments. Later, they \cite{cdc_shou_cvpr17} refined this work by designing a convolution de-convolution network to generate dense predictions for refining action boundaries. Zhao et al. \cite{SSN2017ICCV} decomposed their model into separate classifiers for classifying actions and determining completeness. Meanwhile, due to success in object detection for localizing objects in images, recent works started applying similar ideas to videos. Xu et al. \cite{r-c3d} presented a R-CNN like action detection framework from C3D features. Dai et al. \cite{Dai_2017_ICCV} argued for the importance of sampling at two temporal scales to capture temporal context in a faster R-CNN architecture.  

Our model is also capable of performing action detection, as we generate a dense labeling of videos. When combined with pre-trained action proposals, our model can be used for classifying these proposals (by average pooling the per-frame classification score computed by TAN and multiplying it with the actionness score of the proposal), which demonstrates that it can also be useful in other video recognition tasks.

\section{Architecture}
In this section, we introduce our Temporal Aggregation Network (TAN) in detail. We summarize existing methods and describe the difference. A detailed analysis and intuition of our key building blocks is then provided.

\subsection{Temporal Modeling}
Current approaches to temporal modeling generally fall in two categories:


\textbf{2D Convolution + Late fusion.} Many previous works treat video as a collection of frames. A common approach is to extract frame-by-frame features from deep networks pre-trained on the ImageNet classification task. Subsequently, these frame-wise features are fused using variants of pooling, temporal convolutions or LSTMs. To compensate for missing motion cues from static image features, a second network extracts motion features from flow or improved dense trajectories (IDT) separately - commonly known as two-stream approaches.

\textbf{3D Convolution.} Alternatively, 3D convolutions can be used to model video directly. 3D convolutions are a natural extension to traditional 2D convolutions with an extra dimension spanning the temporal domain. 3D convolution networks form a hierarchical representation of spatio-temporal data. However, because of an additional dimension in the convolution kernels, 3D convolutional networks have significantly more parameters compared to 2D networks, which leads to difficulty in optimization and over-fitting. 

\textbf{Temporal Aggregation Network.} Our approach differs from both previous approaches. Our goal is to model spatio-temporal information by decomposing 3D convolutions into spatial and temporal dilated convolutions. By stacking both types of convolutions repeatedly, we form a temporal aggregation network (TAN) that not only captures spatio-temporal information but also creates hierarchical representations.

\subsection{Proposed Temporal Aggregation Module}

\begin{figure}[b]
    \begin{center}
        \includegraphics[width=0.95\linewidth]{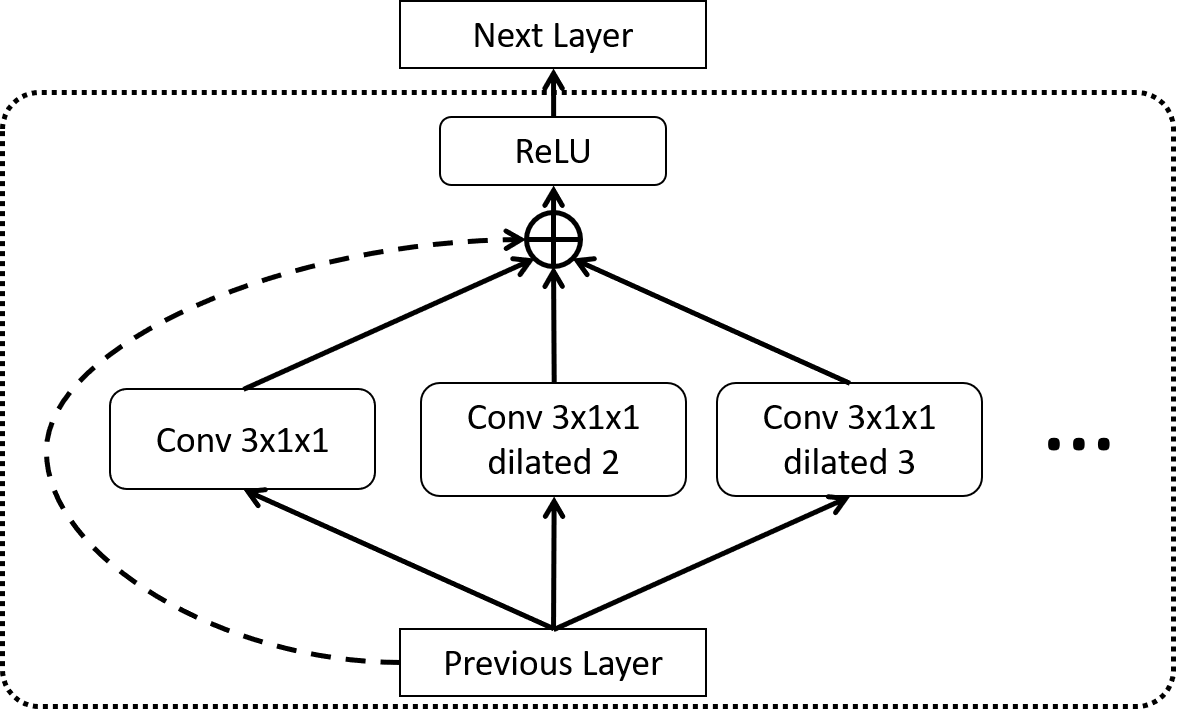}
    \end{center}
    \caption{The proposed temporal aggregation module. It consists of one temporal convolution and several temporal dilated convolutions to capture motions at multiple temporal resolutions.}
	\label{fig:ta}
\end{figure}

A Temporal Aggregation module combines multiple temporal convolutions with different dilation factors (in the temporal domain) and stacks them across the entire network. This supports the capture of context information across multiple spatio-temporal scales. Temporal convolution is a simple 1D convolution. A dilated convolution is a convolution whose filter has been dilated in space by a specific factor. They have been widely used in semantic segmentation and object detection tasks \cite{dilated} to capture spatial context across scales. Here, we apply 1D dilated convolutions in the temporal dimension, see Fig. \ref{fig:ta}. They enable capturing larger context without reducing the resolution of the feature-map. Multiple temporal dilated convolutions not only help to model context across temporal scales but also form an internal attention mechanism to match a large range of motion frequencies.

\textbf{Implementation.} Given any feature inputs from previous layers, we apply one temporal convolution with a filter size of 3 and two dilated temporal convolutions with filter size 3 and dilation factors 2 and 3 respectively. All temporal convolutions are applied across all spatial positions and channels, which help to learn temporal patterns among different filter activations. Inspired by the recent success of residual networks \cite{resnet}, we also add a residual identity connection from previous layers. The responses from all convolutions are further accumulated by conducting element-wise sum  followed by a ReLU non-linear activation, which together behave as a soft attention based weighted selection among different temporal resolutions. The proposed module is illustrated in Figure \ref{fig:ta}.

\textbf{Relation to LSTM.} Long short term memory (LSTM) units are normally used in sequence modeling. By using a gated connection, LSTM allows a network to "remember" important context clues over time. Our temporal aggregation network behaves similarly by modeling temporal information using temporal convolutions. Additionally, our module is capable of distinguishing motion frequency differences among temporal resolutions, which is particularly useful when applied to action recognition tasks where actions having varying lengths. 

\textbf{Relation to Wavenet.} Our temporal aggregation module is partially inspired by the Wavenet \cite{wavenet} model. Our module also uses dilated convolutions to model different temporal context at many time-scales. But there are significant differences. The major one is that our module is used in conjunction with spatial convolutions to capture both low-level and high-level temporal consistency among spatial positions, which is critical for modeling spatio-temporal relationships. Additionally, since we apply our module across different channels of spatial filter responses, it also behaves as a object instance trajectory learner similar to \cite{WangQT15a}.

\textbf{Relation to stack of convolutions with different filter sizes.} It is possible to model multiple temporal resolutions by just stacking convolutions with different filter sizes. However, one of the main reasons we choose dilated convolution over convolution is computing cost. Multi-scale convolutions such as those used in models like inception \cite{inception}, have to largely increase the size of learnable filters. To accomplish effective learning, one usually needs to resort to dimensionality reduction as the response channels increase in size rapidly. By using dilated convolution, we are able to achieve the same goal without reducing temporal resolution. 

\subsection{Spatial and Temporal Convolution Stacking}
As discussed previously, we repeatedly stack spatial and temporal convolutions to model spatio-temporal information and create a hierarchical spatio-temporal representation. We need a deep architecture that can be effectively trained. Multi-entity actions, like ``person picking a book", typically involve larger and semantically richer spatio-temporal motion patterns which are better represented in deeper layers. On the other hand, other actions may be specific to an object instance, like ``person smiling" which require high resolution spatio-temporal features (for smiling) while still understanding deeper semantic concepts like ``person". We seek to capture high-level semantic concepts while preserving high-resolution features generated in the early layers of the network.
Residual networks are a good candidate since their linear structure with residual connections preserves information learned across different convolution levels.

To model spatial information, we borrow the bottleneck structure from residual networks. A bottleneck block is a stack of 3 convolution layers with filter size 1x1, 3x3 and 1x1. The 1x1 convolution layers are used to adjust dimensions and the 3x3 convolution layer behaves as a bottleneck with smaller input/output dimensions. A residual identity connection is also attached to accelerate learning. Each bottleneck block is followed by a ReLU non-linear activation.  

\begin{figure*}
    \begin{center}
        \includegraphics[width=1\linewidth]{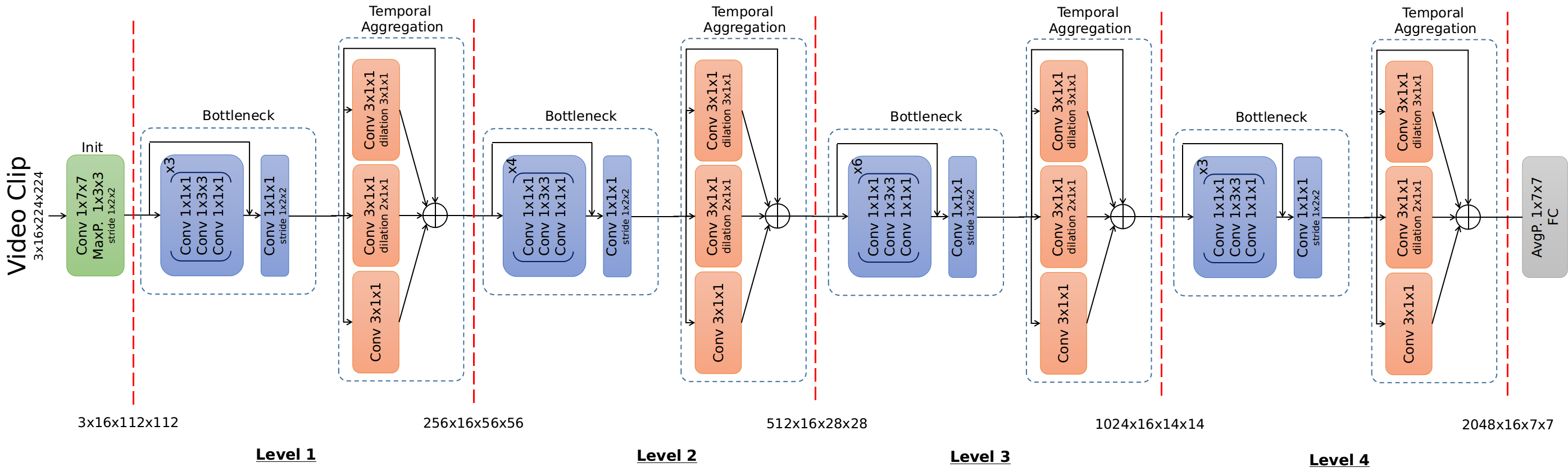}
    \end{center}
    \caption{A closer look of our network architecture. Our model contains four levels. Each level consists of several bottleneck blocks and one temporal aggregation module at the end. The spatial resolution is reduced by two after each level, while the temporal resolution is not reduced. }
\label{fig:arch}
\end{figure*}

The intuition behind repeatedly stacking spatial and temporal blocks among different 
levels is straight-forward. In the early stage of the network, we capture small spatio-temporal patterns using temporal convolutions as the spatial receptive field is small. In the deeper stages of the network, we progressively capture larger spatio-temporal patterns using temporal convolutions as the spatial receptive field increases. Meanwhile, our temporal aggregation blocks support capturing both fast and slow actions across temporal scales and across spatial levels. In this way, we form hierarchical spatio-temporal representations.

Stacking filters in the same block was first proposed in the VGG network \cite{vgg} in which a single large convolution of size 7x7 was broken down into three successive kernels. Such a stacking of convolution kernels increased the depth of the network and also reduced the model complexity (as three 3x3 kernels have fewer parameters than one 7x7 kernel). The receptive field of the network was also the same as the original network which used 7x7 convolutions. Since neural networks are prone to over-fitting, by reducing the model complexity while preserving the desirable characteristics of the network (Zeiler and Fergus \cite{zeiler2014visualizing}), the VGG network reduced over-fitting and hence was able to improve its performance significantly on the image classification task. Our approach of stacking temporal convolutions after spatial convolutions employs a similar strategy for reducing model complexity - albeit for modelling spatio-temporal patterns, where over-fitting is a serious concern.

\subsection{Full Model}
The final architecture consists of four levels of bottleneck blocks for spatial modeling and temporal aggregation blocks for temporal modeling. In each level, there are multiple bottleneck blocks following by one temporal aggregation block similar to a residual network. 

The weights of bottleneck blocks can be initialized using pre-trained ImageNet models. The spatial resolution is reduced after the initial convolution and pooling layers and after every level with max pooling. Notice that there is no temporal resolution reduction due to dilated convolutions in our temporal aggregation blocks. After all four levels of feature learning, the final feature outputs are pooled by average pooling and mapped to semantic space by a fully connected layer. Depending on the specific task, our network can both output an aggregated prediction or dense predictions with no extra cost. Our full model is illustrated in Figure \ref{fig:arch}.

\section{Experiments}
We conducted a series of ablation studies to illustrate the efficiency and effectiveness of our model. We then compare to state-of-the-art methods on dense multi-label action recognition tasks.

\subsection{Datasets} 

Charades \cite{charades} contains 157 action classes from common everyday activities collected by 267 people at home. It has 7986 untrimmed videos for training and 1864 untrimmed videos for validation with an average length of 30 seconds. It is considered as one of the most challenging multi-label action recognition dataset. 

MultiTHUMOS \cite{multithumos} contains 65 action classes and another 413 videos apart from the original THUMOS datatset \cite{THUMOS14} and extends it to a dense multi-label task with 1.5 labels per frame and 10.5 distinct action categories per video. It is suitable for an in-depth study of simultaneous human actions in video. 

THUMOS14 \cite{THUMOS14} contains 20 sport action classes from over 24 hours of videos. The detection task contains 2765 trimmed videos for training, 1,010 untrimmed videos for validation, but only 200 untrimmed videos in validation and 213 untrimmed videos in testing set have foreground labels. This dataset is quite challenging because it contains long videos of multiple short action instances.

\begin{table}
    \begin{center}
        \begin{tabular}{l|c|c}
        \hline
        Method  & ~ Frame mAP ~ & ~ Video mAP ~ \\
        \hline  
        Res50-2D  & 11.7 & 21.1\\
        Res50-3D  & 10.3 & 19.0 \\
        \hline
        Res50-TAN (Ours) ~ & 14.2 & 25.5\\
        \hline
        \end{tabular}
    \end{center}
    \caption{Comparison of different temporal modeling architectures on Charades.}
    \label{tb:arch}
\end{table}

\begin{table*}[h]
	\centering
	\begin{tabular}{c|c|c|c|c|c|c}
		\hline
		\multicolumn{5}{c|}{Components}  & ~ Frame mAP ~ & ~ Video mAP ~ \\
		\multicolumn{1}{c|}{Lv. 4} & \multicolumn{1}{c|}{Lv. 3} & \multicolumn{1}{c|}{Lv. 2} & \multicolumn{1}{|c}{Lv. 1} & \multicolumn{1}{|c}{Dilation} & \multicolumn{1}{|c}{} & \multicolumn{1}{|c}{}\\
		\hline
		\checkmark & $\times$  & $\times$  & $\times$ & \checkmark  & 12.2 &22.1 \\
		\checkmark & \checkmark & $\times$ & $\times$ & \checkmark  & 13.1 &23.6 \\
		\checkmark & \checkmark & \checkmark & $\times$ & \checkmark & 13.7 & 24.9 \\
		\checkmark & \checkmark & \checkmark & \checkmark & \checkmark & \textbf{14.2} & \textbf{25.5}  \\
		\checkmark & \checkmark & \checkmark & \checkmark & $\times$ & 12.2 &22.4 \\
		\hline
		
	\end{tabular}
	\caption{Ablation study for placing temporal aggregation module at different levels and the effectiveness of dilation}
	\label{tb:abl}
\end{table*}

\subsection{Ablation Study}
\textbf{Analysis of Temporal Modeling Methods.} We first compare our model with two baselines. The three models are:
\begin{itemize}
  \item \textbf{Res50-2D}: A model based on ResNet-50, trained frame-by-frame and with an LSTM at the end to generate predictions.
  \item \textbf{Res50-3D}: A model using 3D convolutions to replace all 2D convolutions in ResNet-50. For example, we use 3x3x3 convolutions instead of 3x3 convolutions and use a stride size of 2x2x2 instead of 2x2 when needed. 
  \item \textbf{Res50-TAN}: Our proposed method with spatial convolution layers initialized from ResNet-50.
\end{itemize}
All models take a 16 frame clip with 224x224 resolution as input in 4 FPS, which corresponds to about 4 seconds. As 3D convolutions require large number of training data, to maintain fairness in experiments, all models are pre-trained on the recently released large scale Kinetics dataset \cite{i3d} and then fine-tuned and tested on the Charades dataset \cite{charades}. The results are shown in Table \ref{tb:arch}. We evaluated on both frame-level mAP and video-level mAP.  Our proposed TAN clearly outperforms the baseline methods, while the 3D convolution version performs worst due to the reduction of temporal resolution. 

\textbf{Effectiveness of the Temporal Aggregation Module.} We evaluate the effectiveness of our network architecture by conducting an ablation study by varying the position of the temporal aggregation module in the network, such as only at the final level 4, only at level 3,4 and at level 2,3,4. We also describe an experiment by replacing our temporal aggregation module with a simple 3x1x1 temporal convolution. Table \ref{tb:abl} shows the comparative results. As more temporal aggregation modules are added, performance improves. This shows that our model can learn hierarchical spatio-temporal patterns. Meanwhile, we also evaluate the effectiveness of temporal dilated convolutions in the temporal aggregation module. When we replace our temporal aggregation module with a simple temporal convolution, performance drops significantly by 2\% on Frame mAP and 3.1\% on Video mAP. This shows that our temporal aggregation module is more effective at modeling temporal information from multiple temporal scales. 

\textbf{Impact of Sampling Rate in Videos.}  
We train and test our model with four different video sampling rates of 1,2,4,8 FPS. As seen in Table \ref{tb:st}, the performance doesn't drop significantly until we aggressively drop the video sampling rate to 1FPS. Our model shows reasonable immunity to variances of temporal resolution.

\begin{table}
    \small
    \begin{center}
        \begin{tabular}{l|c|c|c|c}
        \hline
        \quad Sampling rates \quad& ~ 1 FPS ~ & ~ 2 FPS ~ & ~ 4 FPS ~ & ~ 8 FPS ~ \\
        \hline
        \quad Video mAP & 21.3  & 25.3  & 25.5  & 24.1\\
        \hline
        \end{tabular}
    \end{center}
    \caption{Comparison of performance using different temporal stride on Charades}
    \label{tb:st}
\end{table}

\textbf{Impact of Networks Layers.} We also conduct an ablation study to investigate the influence of depth of networks. Here, we only vary the number of bottleneck blocks used in level 3 and level 4, similar to ResNet, leaving the temporal blocks unchanged at the end of each level. Table \ref{tb:layer} shows that our model can further benefit from deeper networks.

\begin{table}
    \small
    \begin{center}
        \begin{tabular}{l|c|c|c}
        \hline
         Number of layers & $50_s+12_t$ & $101_s+12_t$ & $152_s+12_t$\\
        \hline
         Frame mAP & 14.2 & 15.5  & 17.6 \\
         Video mAP & 25.5 & 27.5  & 31.0 \\
        \hline
        \end{tabular}
    \end{center}
    \caption{Comparison of performance using different depth of spatial convolutions on Charades}
    \label{tb:layer}
\end{table}

\textbf{Impact of Pretraining.} Finally, we conduct an ablation study to investigate the influence of pretraining. Table \ref{tb:ptr} shows that our model can further benefit from larger dataset, meanwhile even without pretraining, our method still achieves state-of-the-art performance compared to previous approaches.

\begin{table}
    \small
	\begin{center}
		\begin{tabular}{l|c|c}
			\hline
			Method & Frame mAP & Video mAP \\
			\hline
			Ours w/o pretraining & 14.0 & 25.3\\
			Ours w/ pretraining& 17.6 & 31.0\\
			\hline
		\end{tabular}
	\end{center}
	\caption{Comparison of performance with and without Kinetics pretraining on Charades}
	\label{tb:ptr}
\end{table}

\subsection{Multi-label Action Recognition}
 We conduct experiments on two popular multi-label action recognition datasets and compare our results to state-of-the-art methods.

\textbf{Implementation details.} For all datasets, we use a temporal aggregation network based on ResNet-152 with a 16 frame temporal resolution. It is pre-trained on the Kinetics dataset and fine-tuned later. For the  Charades dataset, we use a sampling rate of 4 FPS. We optimized our model using the Adam optimizer with a learning rate of $10^{-4}$ for the first 10 epochs and $10^{-5}$ in the following 10 epochs. It is evaluated using two metrics following \cite{charades}: one is a video level mAP for video based multi-label classification and the other is a frame level mAP uniformly extracted at 25 frames per video to approximate the multi-label action detection task. For video level classification, average pooling is used to map dense frame level labels to video level labels. For MultiTHUMOS, we use a sampling rate of 10 FPS following \cite{multithumos}. We also optimized our model using the Adam optimizer with a learning rate of $10^{-4}$ for the first 10 epochs and $10^{-5}$ in the following 5 epochs. Finally, MultiTHUMOS is also evaluated using the standard frame level mAP metric.

\begin{table}
	\begin{center}
		\begin{tabular}{l|c|c}
			\hline
			Method & Frame mAP & Video mAP \\
			\hline
			Random \cite{charades} & -- & 5.9\\
			C3D \cite{c3d} & -- & 10.9\\
			IDT \cite{idt} & -- & 17.2\\
			Two-stream \cite{feichtenhofer2016convolutional} & 8.9 & 14.3\\
			Asynchronous Temp Field \cite{sigurdsson2017asynchronous} & 12.8 & 22.4\\
			ActionVLAD \cite{Girdhar_ActionVLAD_cvpr17} & -- & 21.0\\
			Predictive-corrective \cite{predictive_cvpr17} & 8.9 & --\\
			R-C3D \cite{r-c3d} & 12.7 & --\\
			\hline
			\textbf{Ours} & \textbf{17.6} & \textbf{31.0}\\
			\hline
		\end{tabular}
	\end{center}
	\caption{Comparison with state-of-the-art methods on Charades}
	\label{tb:c1}
\end{table}

\begin{table}
    \begin{center}
        \begin{tabular}{l|c}
        \hline
        Method & Frame mAP \\
        \hline
        IDT \cite{idt} & 13.3\\
        Single-frame CNN \cite{vgg} & 25.4 \\
        Two-Stream \cite{feichtenhofer2016convolutional}& 27.6\\
        Multi-LSTM \cite{multithumos}& 29.6\\
        Predictive-corrective \cite{predictive_cvpr17}& 29.7\\
        \hline
        \textbf{Ours} & \textbf{33.3}\\
        \hline
        \end{tabular}
    \end{center}
    \caption{Comparison with state-of-the-art methods on MultiTHUMOS using frame level mAP}
    \label{tb:mt}
\end{table}

\textbf{Comparison with state-of-the-art.} On Charades, we compare with several state-of-the-art methods such as \cite{charades,c3d,idt,feichtenhofer2016convolutional,sigurdsson2017asynchronous,Girdhar_ActionVLAD_cvpr17,predictive_cvpr17}. Our results are shown in Table \ref{tb:c1}. We first show results of our model without Kinetics pre-training, and it still outperforms state-of-the-art methods by 1.2\% on frame level mAP and 2.9\% on video level mAP.
Our model with Kinetics pre-training obtains a significant improvement over other methods and outperforms them by 4.8\% on frame level mAP and 8.6\% on video level mAP respectively. On MultiTHMOS, we compare with \cite{vgg,feichtenhofer2016convolutional,multithumos,predictive_cvpr17} in Table \ref{tb:mt}. Our model performs 3.6\% better than the previous state-of-the-art method \cite{predictive_cvpr17}.
We also compare our model with several recent network on THUMOS14, such as \cite{vgg,feichtenhofer2016convolutional,multithumos,cdc_shou_cvpr17,predictive_cvpr17}. As seen in Table \ref{tb:det}, our model again outperforms previous state-of-the-art methods \cite{cdc_shou_cvpr17} by 2.4\% on frame-level evaluation.

\begin{table}
	\begin{center}
		\begin{tabular}{l|c}
			\hline
			Method & Frame mAP \\
			\hline
			Single-frame CNN \cite{vgg} & 34.7 \\
			Two-Stream \cite{feichtenhofer2016convolutional}& 36.2\\
			Multi-LSTM \cite{multithumos}& 41.3\\
			Predictive-corrective \cite{predictive_cvpr17}& 38.9\\
			CDC \cite{cdc_shou_cvpr17} & 44.4 \\
			\hline
			\textbf{Ours} & \textbf{46.8}\\
			\hline
		\end{tabular}
	\end{center}
	\caption{Comparison with state-of-the-art methods on THUMOS14 using frame level mAP}
	\label{tb:det}
\end{table}

\begin{figure}
	\begin{center}
		\includegraphics[width=0.95\linewidth]{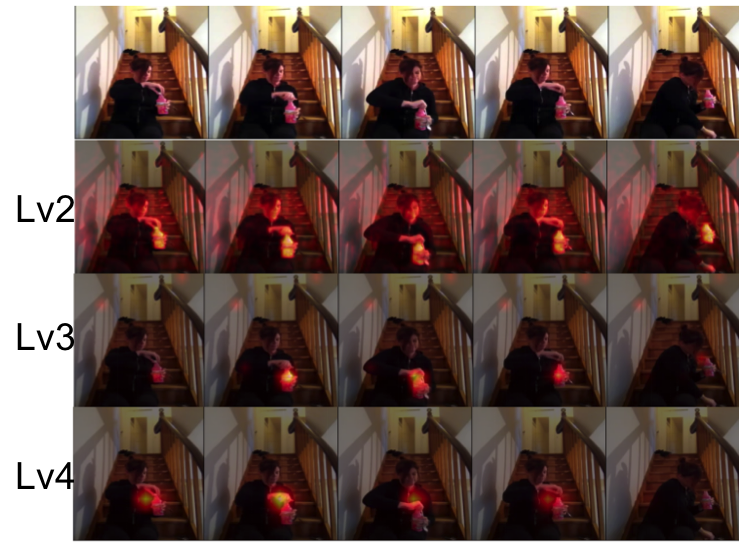}
	\end{center}
	\caption{Visualization of network filter responses from multiple levels. From top to bottom row, each row shows filter responses of a layer from lower to higher layer.
	}
	\label{fig:viz}
\end{figure}

\begin{figure*}[!hbt]
	\begin{center}
		\includegraphics[width=1\linewidth]{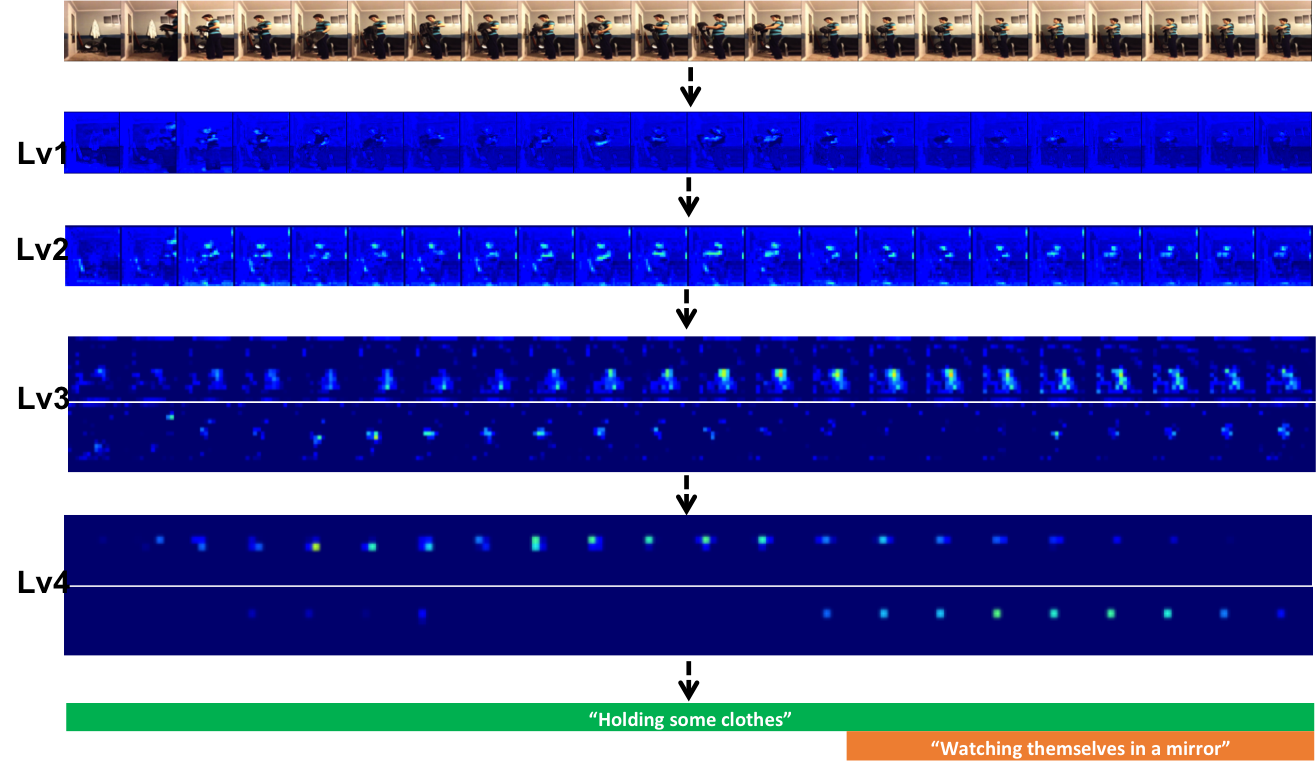}
	\end{center}
	\caption{Visualization of representative filter responses after each level of temporal aggregation block.}
	\label{fig:layers}
\end{figure*}

\begin{figure*}[!hbt]
	\begin{center}
		\includegraphics[width=1\linewidth]{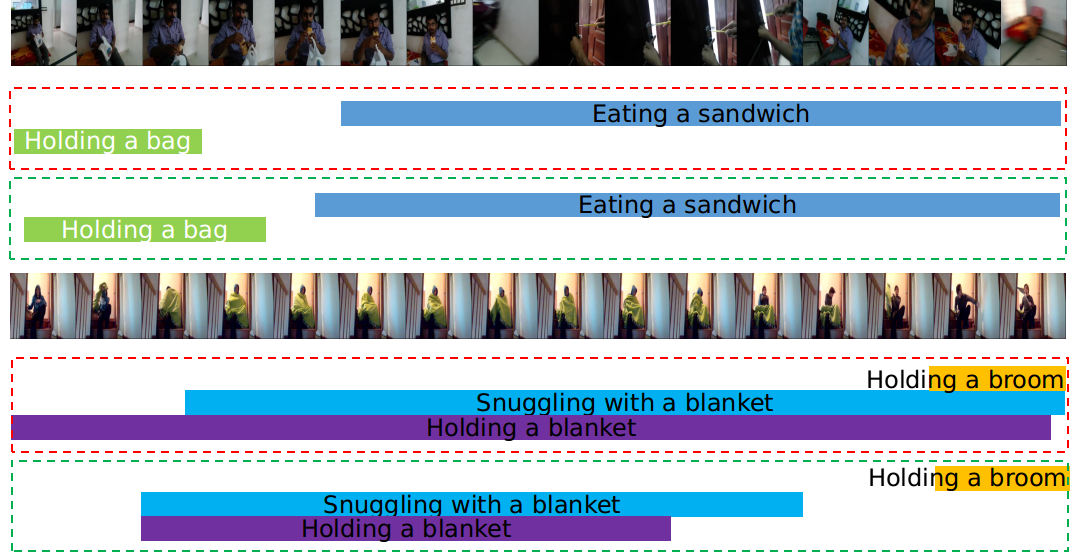}
	\end{center}
	\caption{Qualitative results on the Charades dataset. Red and green dashed boxes show the ground-truth and our predictions respectively. Our network is able to correctly localize multiple actions of variable durations simultaneously.}
	\label{fig:qua}
\end{figure*}

\subsection{Visualization}
We visualize the filter responses of our network to better understand what is learned in different levels in Figure \ref{fig:viz}. In the example, we observe that the early layer filters have responses spread over a larger part of the image, while the later layers learn to focus where the action is present. To better understand how the network learns spatial-temporal consistency, we also visualize representative filter responses after each level of temporal aggregation block, shown in Figure \ref{fig:layers}. In spatial domain, the sallow layers tend to highlight fine-grained movement (such as facial, hand movement) and the deep layers tend to highlight large movement (such as body movement), due to the spatial effective receptive window increases as network goes deep. In temporal domain, the network is capable of capturing action sequence of variable length, thanks to our temporal aggregation blocks. Meanwhile, it is interesting to see that how spatial-temporal responses group up as the network advances. This demonstrates that our model progressively constructs hierarchical spatio-temporal representations for recognizing human actions. 

We also present qualitative results for different samples in Figure \ref{fig:qua}. These results show that our network is effective in localizing both long and short duration actions simultaneously. Notice that it also predicts labels of multiple actions happening at the same time instance.

\section{Conclusion}
We presented a temporal aggregation module that combines temporal convolution with varying levels of dilated temporal convolutions to capture spatio-temporal information across multiple scales. We compare our design with multiple temporal modeling methods. Based on this, we designed a deep network architecture for video applications. Our model stacks spatial convolutions and the new temporal aggregation module repeatedly to learn a hierarchical spatio-temporal representation. It is both efficient and effective compared to existing methods such as 3D convolution networks and two-stream networks. We conduct ablation studies to analyze our model and experiments on multi-label action recognition and pre-frame action recognition datasets prove the effectiveness of our method. 

{\small
\bibliographystyle{ieee}
\bibliography{egbib}
}

\end{document}